\documentclass[11pt]{article}
\usepackage[nohyperref]{jmlr2e}

\usepackage[utf8]{inputenc} %
\usepackage[T1]{fontenc}    %
\usepackage{hyperref}       %
\usepackage{xcolor}
\hypersetup{
  colorlinks,
  linkcolor={red!50!black},
  citecolor={blue!50!black},
  urlcolor={blue!80!black}
  }
  \definecolor{orange}{HTML}{ff7f0e}
  \definecolor{blue}{HTML}{1f77b4}
  
\usepackage{url}            %
\usepackage{booktabs}       %
\usepackage{amsfonts}       %
\usepackage{nicefrac}       %
\usepackage{microtype}      %
\usepackage{comment}     
\usepackage{amssymb}
\usepackage{amsmath,amsfonts,bm}
\usepackage{nicefrac}
\usepackage{textcomp}
\usepackage[capitalise]{cleveref}
\usepackage{enumitem}%
\usepackage{mathtools}
\usepackage{todonotes}

\title{A Comprehensive Survey on Knowledge Distillation of Diffusion Models}

\author{%
}

\author{%
  \name Weijian Luo \email luoweijian@stu.pku.edu.cn\\
   \addr School of Mathematical Sciences\\
   Peking University\\
   Beijing, 100871, China\\\\
}

\editor{}

\def\eqref#1{equation~\ref{#1}}

\def\1{\bm{1}}

\DeclareMathAlphabet{\mathsfit}{\encodingdefault}{\sfdefault}{m}{sl}
\SetMathAlphabet{\mathsfit}{bold}{\encodingdefault}{\sfdefault}{bx}{n}

\usepackage[framemethod=TikZ]{mdframed}

\makeatletter
\usepackage{xspace} 
\def\@onedot{\ifx\@let@token.\else.\null\fi\xspace}
\DeclareRobustCommand\onedot{\futurelet\@let@token\@onedot}

\usepackage{amsmath}

\begin{document}

\maketitle

\begin{abstract}
Diffusion Models (DMs), also referred to as score-based diffusion models, utilize neural networks to specify score functions. Unlike most other probabilistic models, DMs directly model the score functions, which makes them more flexible to parametrize and potentially highly expressive for probabilistic modeling. DMs can learn fine-grained knowledge, i.e., marginal score functions, of the underlying distribution. Therefore, a crucial research direction is to explore how to distill the knowledge of DMs and fully utilize their potential. Our objective is to provide a comprehensible overview of the modern approaches for distilling DMs, starting with an introduction to DMs and a discussion of the challenges involved in distilling them into neural vector fields. We also provide an overview of the existing works on distilling DMs into both stochastic and deterministic implicit generators. Finally, we review the accelerated diffusion sampling algorithms as a training-free method for distillation. Our tutorial is intended for individuals with a basic understanding of generative models who wish to apply DM's distillation or embark on a research project in this field.
\end{abstract}

\section{Introduction}
\subsection{ARMs: Model Explicit Likelihood Functions}
The objective of deep generative modeling is to train neural-parametrized models that can generate highly realistic data samples. Numerous deep generative models have been proposed to achieve this objective, each from a different perspective. In general, generative models aim to express and approximate certain sufficient characteristics of the underlying data distribution by minimizing probability divergences or metrics. The likelihood function, which is the log density of the underlying distribution, is a commonly used distributional characteristic. Auto-regressive models (ARMs) \citep{graves2013generating, van2016pixel, jozefowicz2016exploring} are representative models that use neural networks to parametrize the log-likelihood function, i.e., the logarithm of probability density functions, and learn to match the underlying data's log-likelihood. ARMs are trained using KL divergence minimization. Let $p_d$ denote the underlying data distribution, from which we only have access to consistent samples, i.e., $x\sim p_d$. ARMs sum up a sequence of outputs of neural networks with strict orders to explicitly express the conditional factorization of the model's likelihood function which writes
\begin{align}\label{eqn:arm_logp}
    p_\theta(x) = f_\theta(x^{(1)})+ \Sigma_{i=2}^D f_\theta(x^{(i)}|x^{(1)},...,x^{(i-1)}).
\end{align}
Here $x^{(i)}$ represents the $i$-th coordinates of a point $x\in \mathbb{R}^D$ following some strict order. If $x$ is continuous and in a Euclidean space, the term $f_\theta(x^{(i)}|x^{(1)},...,x^{(i-1)})$ is often implemented via the conditional distribution of Gaussian distribution with mean and variance obtained from outputs of neural networks as 
\begin{align}\label{eqn:arms_cont_cond}
    f_\theta(x^{(i)}|x^{(1)},...,x^{(i-1)}) = \log \mathcal{N}(x^{(i)}; \mu_\theta(x^{(1)},..,x^{(i-1)}), \Sigma_\theta (x^{(1)},..,x^{(i-1)})).
\end{align}
Here $\mathcal{N}(x;\mu,\Sigma)$ represents the density functions of multivariate distribution with mean $\mu$ and covariance matrix $\Sigma$. The terms $\mu^{(i)}_\theta$ and $\Sigma^{(i)}_\theta$ are usually implemented as the outputs of some neural networks. For discrete data, the conditional distribution in \eqref{eqn:arm_logp} is usually implemented via a soft-max output of some neural networks with the expression
\begin{align}\label{eqn:arms_cont_discrete}
    f_\theta(x^{(i)}|x^{(1)},...,x^{(i-1)}) = \operatorname{Softmax} (g_\theta(x^{(1)},..,x^{(i-1)})).
\end{align}
Here $g_\theta(.)$ is a neural network whose output has the same dimension as $x^{(i)}$'s discrete range. 

 The KL divergence between $p_d$ and $p_\theta$ is defined as
\begin{align}\label{eqn:kl_def}
    \mathcal{D}_{KL}(p_d,p_\theta) \coloneqq \mathbb{E}_{x\sim p_d} \big[ \log p_d(x) - \log p_\theta(x) \big]= \mathbb{E}_{p_d}\big[\log p_d(x) \big] - \mathbb{E}_{p_d} \big[\log p_\theta(x)\big].
\end{align}
Since the first term of \eqref{eqn:kl_def} does not depend on parameters $\theta$, training ARMs by minimizing the KL divergence between $p_d$ and $p_\theta$ is equivalent to 
\begin{align}\label{eqn:mle_def}
    \theta^* = \arg\min_\theta \mathcal{D}_{KL}(p_d,p_\theta) = \arg\max_\theta \mathbb{E}_{p_d} \big[\log p_\theta(x) \big].
\end{align}
The optimal $\theta^*$ in problem \eqref{eqn:mle_def} is also called a maximum likelihood estimation (MLE) of the parameter $\theta$. The term $\mathcal{L}(\theta) = \mathbb{E}_{p_d} \big[\log p_\theta(x) \big]$ is the expected likelihood. Two important points about ARMs need to be emphasized. Firstly, ARMs explicitly model log-likelihood functions (\eqref{eqn:arms_cont_cond} and \eqref{eqn:arms_cont_discrete}), which limits the implementation's flexibility. Secondly, ARMs require a strict sequential order in their generating algorithm, which makes sampling from ARMs computationally inefficient.

\subsection{EBMs: From Explicit Likelihoods To Unnormalized Ones}
ARMs use neural networks to directly express normalized likelihood functions through the conditional factorization formula. However, this normalization is too limiting to unleash the full potential of neural networks. For instance, the implementation \eqref{eqn:arms_cont_cond} of continuous-valued ARMs constrains the conditional distribution to be a multivariate Gaussian distribution, which may not hold for real-world data distributions. This constraint prevents ARMs from fully matching the data distribution even if the neural networks have infinite expressive capacity. Energy-based models (EBMs) \citep{LeCun2006ATO, Zhu1998FiltersRF}, on the other hand, overcome this normalization issue by using unconstrained neural networks to express and match the data's potential functions, i.e., the logarithm of the unnormalized density functions. Formally, the potential function of a distribution $p$ is defined as a family of functions that satisfy the equation
\begin{align}
    e^{E(x)}/\int e^{E(x)}dx = p(x).
\end{align}
Intuitively, the potential function $E(x)$ in EBMs represents the logarithm of the unnormalized part of the density $p(x)$. Since the potential function does not need to be normalized itself, EBMs use neural networks $E_\theta$ to express model potential functions. The most basic training strategy of EBMs is to also use the maximum likelihood training \eqref{eqn:mle_def}. Specifically, since the EBM-induced distribution has the form $p_\theta(x) = e^{E_\theta(x)}/Z_\theta$, where $Z_\theta = \int e^{E_\theta(x)}dx$ is the normalizing constant, which is often challenging to evaluate and is viewed as intractable in most cases. The expected likelihood of EBM-induced distribution writes
\begin{align}\label{eqn:ebm_logp}
    \mathcal{L}(\theta) = \mathbb{E}_{p_d} \big[\log p_\theta(x) \big] = \mathbb{E}_{p_d} \big[E_\theta(x)\big] - \log Z_\theta.
\end{align}
Taking the parameter gradient of expected likelihood \eqref{eqn:ebm_logp}, we have
\begin{align}\label{eqn:ebm_logp_grad}
    \frac{\partial}{\partial\theta}\mathcal{L}(\theta) &= \mathbb{E}_{p_d} \big[\frac{\partial}{\partial\theta}E_\theta(x)\big] - \frac{\partial}{\partial\theta} \log Z_\theta\\
    &= \mathbb{E}_{p_d} \big[\frac{\partial}{\partial\theta}E_\theta(x)\big] - \frac{1}{Z_\theta} \frac{\partial}{\partial\theta} Z_\theta\\
    &= \mathbb{E}_{p_d} \big[\frac{\partial}{\partial\theta}E_\theta(x)\big] - \frac{1}{Z_\theta} \frac{\partial}{\partial\theta} \int e^{E_\theta(x)}dx\\
    &= \mathbb{E}_{p_d} \big[\frac{\partial}{\partial\theta}E_\theta(x)\big] - \frac{1}{Z_\theta} \int \big[ \frac{\partial}{\partial\theta} E_\theta(x)\big] e^{E_\theta(x)}dx\\
    &= \mathbb{E}_{p_d} \big[\frac{\partial}{\partial\theta}E_\theta(x)\big] - \int \big[ \frac{\partial}{\partial\theta} E_\theta(x)\big] \frac{e^{E_\theta(x)}}{Z_\theta}dx\\
    &= \mathbb{E}_{p_d} \big[\frac{\partial}{\partial\theta}E_\theta(x)\big] - \mathbb{E}_{p_\theta} \big[ \frac{\partial}{\partial\theta} E_\theta(x)\big].
\end{align}
To calculate the expected log-likelihood gradient with respect to the parameter $\theta$, we need to obtain consistent samples from the EBM-induced distribution $x \sim p_\theta$, which is an un-normalized distribution. Fortunately, several MCMC algorithms are capable of generating such samples, including those proposed by \citet{robert1999monte}, \citet{hastings1970monte}, \citet{roberts1998optimal}, \citet{xifara2014langevin}, and \citet{neal2011mcmc}. By combining the gradient formula \eqref{eqn:ebm_logp_grad} with MCMC algorithms, it is possible to train EBMs using maximum likelihood estimation. The key distinction between energy-based models (EBMs) and auto-regressive models (ARMs) lies in how they utilize neural networks. While EBMs employ unconstrained neural networks to model potential functions, ARMs use neural networks as a component of explicit conditional densities. This difference in approach allows EBMs to tap into the full expressive power of neural networks by avoiding the constraints imposed by normalization requirements.

\subsection{SBMs and DMs: From Potentials to Scores} 
In the preceding sections, we introduced EBMs and their training and sampling methods. To approximate data potential functions, EBMs use neural networks and generate samples using MCMC algorithms with learned potentials. Among the various MCMC methods available, Langevin dynamics (LD) or Langevin MC \citep{roberts1996exponential} is a preferred choice for its ease of implementation and good performance even under weak conditions. Let $p(x)$ be a differentiable density function, the LD is defined through a stochastic differential equation,
\begin{align}\label{eqn:ld}
    dX_t = \frac{1}{2}\nabla_{X_t} \log p(X_t)dt + dW_t, ~p^{(0)}=p_0, ~t\in [0,\infty].
\end{align}
Langevin dynamics (LD) is notable for two reasons. Firstly, when $t\to\infty$, the marginal distribution of LD can converge to $p(x)$ regardless of the initial distribution $p_0$, given certain conditions on $p(x)$. Secondly, the simulation of LD only requires the gradient of the potential function, or the score function $\nabla_x \log p(x)$. Hence, even if the distribution $p(x)$ is not normalized, using the potential function $\log p(x)$ for LD still generates valid samples as the normalized distribution. Score-based models (SBMs) take inspiration from LD and utilize neural networks to train a neural score function $S_\theta$ that can match the underlying data distribution. If the neural score function is trained well enough to match the data score functions $S_d(x)\coloneqq\nabla_x \log p_d(x)$, samples from the SBM can be obtained by replacing the score function $\nabla_x \log p(x)$ with the SBM's neural score function $S_\theta(x)$ in the simulation of LD.

 Now, let's turn our attention to the training strategies of SBMs. Since $S_\theta$ directly expresses the model's underlying score functions, the model's potential function $p_\theta$ becomes intractable. Therefore, the commonly used KL divergence for training EBMs and ARMs as explained in previous sections does not work for training SBMs. Instead, SBMs minimize the Fisher divergence, a probability divergence that only requires the model's score functions. Formally, the Fisher divergence between two distributions $p(x)$ and $q(x)$ is defined as 
\begin{align}\label{def:fisher_divergence}
    \mathcal{D}_{F}(p,q) \coloneqq \mathbb{E}_{x\sim p} \|\nabla_x \log p(x) - \nabla_x \log q(x)\|^2_2.
\end{align}
Let $p_d$ denote the data distribution and $S_d(x)=\nabla_x \log p_d(x)$ represents the data score function. The Fisher divergence between data and SBM-induced distribution $p_\theta$ writes
\begin{align}\label{eqn:sbm_fdiv}
    \mathcal{D}_{F}(p_d, p_\theta) &= \mathbb{E}_{p_d} \frac{1}{2}\|S_d(x) - S_\theta(x)\|_2^2 \\
    &= \mathbb{E}_{p_d}\big[ \frac{1}{2}\|S_d(x) \|_2^2 \big] + \mathbb{E}_{p_d}\big[ \frac{1}{2}\|S_\theta(x) \|_2^2 \big] - \mathbb{E}_{p_d}\big[ \langle S_d(x), S_\theta(x) \rangle \big].
\end{align}
In practice, the first term of Fisher divergence \eqref{eqn:sbm_fdiv} does not depend on the parameter $\theta$ and thus can be dropped. The third term is shown to be equivalent to a data-score-free form as
\begin{align*}
    \mathbb{E}_{p_d}\big[ \langle S_d(x), S_\theta(x) \rangle \big] = -\mathbb{E}_{p_d} \sum_{d=1}^D \frac{\partial s_\theta^{(d)}(x)}{\partial x^{(d)}}.
\end{align*}
Here the notation $s_\theta^{(d)}(x)$ denotes the $d$-th component of the model score function and $x^{(d)}$ denotes the $d$-th component of the input vector. So minimizing the Fisher divergence is equivalent to minimizing a tractable objective
\begin{align}\label{eqn:sm_objective}
    \mathcal{L}_{SM}(\theta) \colon = \mathbb{E}_{p_d} \big[\frac{1}{2}\|S_\theta(x)\|_2^2 + \mathbb{E}_{p_d} \sum_{d=1}^D \frac{\partial s_\theta^{(d)}(x)}{\partial x^{(d)}} \big].
\end{align}
The optimization problem in equation \eqref{eqn:sm_objective} is known as Score Matching (SM) \citep{Hyvrinen2005EstimationON}. However, evaluating the gradient term $\langle \nabla_x, S_\theta(x) \rangle$ by taking the data gradient through a neural network can be memory-intensive, which poses challenges when working with high-dimensional data. To overcome this limitation, several approaches have been proposed to improve the efficiency of SM \citep{Song2019SlicedSM, Pang2020EfficientLO, Vincent2011ACB}. Among them, the seminal work \citep{Vincent2011ACB} proposed a so-called denoising score matching (DSM) objective that does not require taking data gradient but instead with the objective
\begin{align}\label{eqn:dsm_0}
    \mathcal{L}_{DSM}(\theta) \colon= \mathbb{E}_{x\sim p_d, \Tilde{x}\sim p(\Tilde{x}|x) } \|S_\theta(\Tilde{x}) - \nabla_{\Tilde{x}} \log p(\Tilde{x}|x)\|^2_2.
\end{align}
In the above paragraph, the objective function $\mathcal{L}_{DSM}(\theta)$ is minimized using a perturbation kernel $p(\Tilde{x}|x)$, which is efficient to sample and has an explicit expression. One common choice for the perturbation kernel is a Gaussian distribution $\mathcal{N}(\Tilde{x};x,\sigma^2 \mathbf{I})$ with a noise variance $\sigma^2$. By minimizing this objective, we can obtain an approximation of the data distribution $\Tilde{p}_d(\Tilde{x})$, which is similar to the original data distribution if the perturbation kernel is not too different from the identity. However, using the LD for sampling can be problematic for high-dimensional data because the data is concentrated around some low-dimensional manifold embedded in high-dimensional space. As a solution, score-based diffusion models use multiple or continuously-indexed perturbation kernels to improve both learning and sampling. This approach can improve the performance of SBMs by generating samples that are more accurate and representative of the original data distribution.

\subsection{Diffusion Models: Multi-Level SBMs}
In contrast to SBMs that use a single score network, score-based diffusion models (DMs) \citep{Song2020ScoreBasedGM} employ a more advanced approach by utilizing a multiple-level or continuous-indexed score network $S_\theta(x,t)$. Additionally, instead of a single perturbation kernel, DMs use a family of conditional transition kernels induced by stochastic differential equations to perturb the data. Consider a forward diffusion SDE
\begin{align}\label{def:forward_sde}
dX_t = F(X_t,t)dt + G(t)dW_t,~X_0\sim p_d^{(0)}=p_d,
\end{align}
where $W_t$ is a Wiener process. Let $p_t(x_t|x_0)$ denote the conditional transition kernel of the forward diffusion \ref{def:forward_sde}, and $p_d^{(t)}$ denote the marginal distribution at diffusion time $t$, initialized with $p_d^{(0)} = p_d$. Two special forward diffusions, variance preserving (VP) diffusion, and variance exploding (VE) diffusion \citep{Song2020ScoreBasedGM} are favored across diffusion model literature. 

\begin{figure}
\centering
\includegraphics[width=0.9\textwidth]{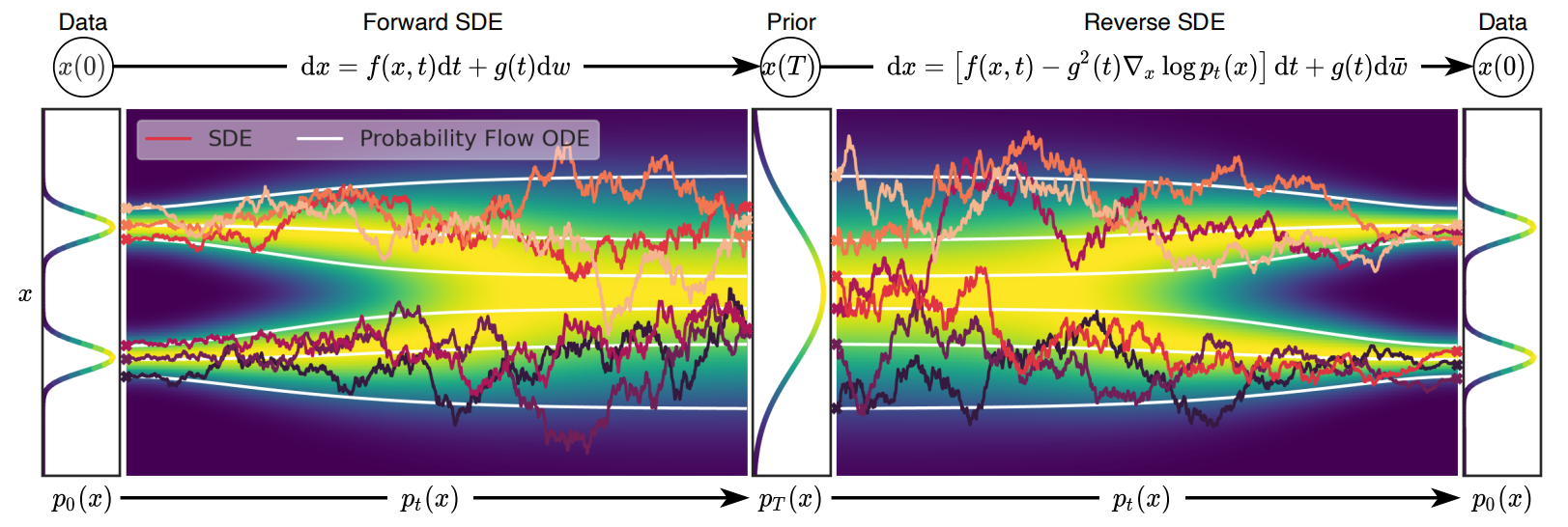}
\caption{Forward and Reversed SDE of Diffusion Models. The figure is taken from \citet{Song2020ScoreBasedGM}.}
\label{fig:dds_sde}
\end{figure}

\paragraph{VP Diffusion} The VP diffusion takes the form of 
\begin{align}\label{eqn:vp_forward}
    dX_t = -\frac{1}{2}\beta(t)X_t dt + \sqrt{\beta(t)}dW_t, ~t\in [0,T],
\end{align}
where $\beta(t)$ is a pre-defined schedule function. The conditional transition kernel of VP diffusion has an explicit expression 
\begin{align}\label{eqn:vp_conditional}
    p(x_t|x_0) = \mathcal{N}(x_t;\sqrt{\alpha_t}x_0;(1-\alpha_t)\mathbf{I}),
\end{align}
where $\alpha_t = e^{-\int_0^t \beta(s)ds}$. With formula \eqref{eqn:vp_conditional}, the simulation of samples of time $t$ is efficient, requiring only a scaling of an initial sample and adding a Gaussian noise
\begin{align}\label{eqn:vp_reparam}
    X_t = \sqrt{\alpha_t} X_0 + \sqrt{1-\alpha_{t}}\epsilon, ~X_0\sim p_0.
\end{align}
$\epsilon\sim \mathcal{N}(0,\mathbf{I})$ is a standard normal vector of the same size as $X_0$. With formula \eqref{eqn:vp_conditional}, obtaining $X_t$ is cheap because we do not need a sequential simulation of the diffusion process. Another advantage of VP diffusion is that under loose conditions, VP diffusion can transport arbitrary initial distribution $p_0$ to a standard multi-variate Gaussian distribution $\mathcal{N}(0, \mathbf{I})$. VP diffusion is perhaps the most widely used forward diffusion across DM literature \citep{}. 
\paragraph{VE Diffusion}
The VE diffusion takes the form 
\begin{align}\label{eqn:ve_forward}
    dX_t = \sqrt{\frac{d \sigma^2(t)}{dt}}dW_t, ~t\in [0, T].
\end{align}
$W_t$ is an independent Weiner process. The transition kernel of VE diffusion writes
\begin{align}\label{eqn:ve_conditional}
    p(x_t|x_0) = \mathcal{N}(x_t; x_0, \sigma(t) \mathbf{I}).
\end{align}
Similar to VP diffusion, the marginal samples of VE diffusion are cheap to be drawn with
\begin{align}\label{eqn:ve_reparam}
    X_t = X_0 + \sigma(t) \epsilon,~X_0\sim p_0
\end{align}
Here $\epsilon\sim \mathcal{N}(0,\mathbf{I})$ is a standard Gaussian vector.

Both VE and VP diffusion processes have been successfully used in diffusion models for various tasks. However, in recent years, several new diffusion processes have been proposed that have either improved the performance of DMs or have been designed for specific tasks. 


\paragraph{Training Method}
DMs minimize a weighted combination of DSM with perturbation kernels $p_t(.|.)$ at each time $t$. More precisely, DMs training objective writes
\begin{align}\label{def:wdsm}
    \mathcal{L}_{WDSM}(\theta) = \int_{t=0}^T w(t) \mathcal{L}_{DSM}^{(t)}(\theta)dt.
\end{align}
\begin{align}\label{eqn:dsm}
\mathcal{L}_{DSM}^{(t)}(\theta) = \mathbb{E}_{x_0\sim p_d^{(0)}, x_t|x_0 \sim p_t(x_t|x_0)} \|S_\theta(x_t,t) - \nabla_{x_t}\log p_t(x_t|x_0)\|_2^2.
\end{align}
By minimizing objective \eqref{def:wdsm}, continuous-indexed score network $S_\theta(x,t)$ is capable of matching marginal score functions of forward diffusion process \eqref{def:forward_sde}. 

In some literature, the DSM objective \eqref{eqn:dsm} is reformulated as a \emph{noise-prediction} objective that trains DMs by learning to predict the added noise. More precisely, consider the VP's transition kernel \eqref{eqn:vp_conditional} as an instance, if $x_t$ and $x_0$ is obtained with re-parametrization technique \eqref{eqn:vp_reparam}, then the gradient terms writes
\begin{align}\label{eqn:vp_cond_grad}
    \nabla_{x_t} \log p_t(x_t|x_0) &= \nabla_{x_t} \bigg[ -\frac{1}{2(1-\alpha_t)} \|x_t - \sqrt{\alpha_t}x_0 \|_2^2 \bigg]\\
    & = \frac{1}{1-\alpha_t} (x_t - \sqrt{\alpha_t} x_0) = \frac{1}{1-\alpha_t} \sqrt{1-\alpha_t}\epsilon\\
    & = \frac{1}{\sqrt{1-\alpha_t}} \epsilon
\end{align}
Combining gradient term \eqref{eqn:vp_cond_grad} and DSM objective \eqref{eqn:dsm}, the DSM objective for VP diffusion can be reformulated as 
\begin{align}\label{eqn:dsm_noise_predict}
    \mathcal{L}_{DSM}(\theta) = & \mathbb{E}_{x_0\sim p_d, \epsilon\sim \mathcal{N}(0,\mathbf{I})} \frac{w(t)}{1-\alpha_t} \|\sqrt{1-\alpha_t}S_\theta(x_t,t) - \epsilon\|_2^2\\
    & = \mathbb{E}_{x_0\sim p_d, \epsilon\sim \mathcal{N}(0,\mathbf{I})} \frac{1}{1-\alpha_t} \|\epsilon_\theta(x_t,t) - \epsilon\|_2^2
\end{align}
The DSM can be reformulated using a noise-prediction objective where a neural network $\epsilon_\theta(x_t,t)$ is trained to predict the added noise $\epsilon$ using the noised data sample $x_t$ and the score network $S_\theta(x_t,t)$. This reformulation involves a modified version of the score network, represented as $\epsilon_\theta(x_t,t) \colon= \sqrt{1-\alpha_t} S_\theta(x_t,t)$, and a re-expression of the diffusion process as $x_t = \sqrt{\alpha_t} x_0 + \sqrt{1-\alpha_t} \epsilon$. A similar formulation can also be applied to the VE diffusion process, which is discussed in detail in the Appendix.

\paragraph{Sampling Strategy} The score network trained in DSM can be utilized in various applications, with one of the most direct applications being the design of a sampling strategy for approximating the underlying data distribution. The fundamental concept behind this mechanism is the existence of a reversed SDE \eqref{def:rev_sde} that has the same marginal distribution as the forward SDE \eqref{def:forward_sde},
\begin{align}\label{def:rev_sde}
    dX_t = [F(X_t,t) - G^2(t)\nabla_{x_t} \log p^{(t)}(X_t)]dt + G(t)d\Bar{W}_t, t\in [T,0], X_T\sim p_d^{(T)}.
\end{align}
Moreover, an ODE \eqref{def:rev_ode} is also found to share the same marginal distribution 
\begin{align}\label{def:rev_ode}
    dX_t = [F(X_t,t) - \frac{1}{2}G^2(t)\nabla_{x_t} \log p^{(t)}(X_t)]dt.
\end{align}
Both the reversed SDE \ref{def:rev_sde} and ODE \ref{def:rev_ode} rely on the true marginal score functions $\nabla_{x_t}\log p_d^{(t)}(x_t)$. However, by replacing the true marginal score functions with the learned neural score functions $S_\theta(x,t)$, generative SDEs and ODEs of DMs can be obtained. Moreover, the concept of generative ODEs can be extended to neural continuous-time normalizing flow models under certain circumstances. By using the learned score functions $S_\theta(x,t)$, sampling from DMs can be achieved through the numerical solutions of sampling SDEs or ODEs. Numerous practical algorithms that use advanced numerical techniques have been developed to improve the generative performance of DMs or enhance the sampling efficiency with minimal loss of performance \citep{song2020denoising,bao2022analytic,liupseudo,zhao2023unipc}.

\paragraph{Successes of DMs} 
Since the pioneering works by \citet{sohl2015deep}, \citet{ho2020denoising}, and \citet{song2020score}, diffusion models (DMs) have emerged as the leading approach for generative modeling, finding widespread use in various domains, including neural image synthesis and editing \citep{nichol2021improved,dhariwal2021diffusion,ramesh2022hierarchical,saharia2022photorealistic,rombach2022high}, audio and molecule synthesis \citep{hoogeboom2022equivariant,chen2020wavegrad}, image segmentation \citep{baranchuk2021label}, and video or 3D object generation \citep{ho2022video,molad2023dreamix,poole2022dreamfusion}. DMs have shown remarkable performance improvements over time, as seen in the steady trend of unconditional Frechet Inception Score \citep{heusel2017gans} (FID) reductions on datasets such as CIFAR10, from 25.32 \citep{song2019generative} to 1.97 \citep{karraselucidating}.

\paragraph{Diffusion Distillation} 
The concept of knowledge distillation \citep{hinton2015distilling, oord2018parallel}, which aims to create smaller and more efficient models while maintaining accuracy, has shown great success in various research domains. In particular, distilling knowledge from pre-trained classifiers has resulted in models with comparable accuracy, reduced model size, and improved inference efficiency \citep{touvron2021training}. Given the success of diffusion models (DMs) in numerous applications, there is a growing interest in distilling knowledge from these models to create smaller and more efficient versions. One of the key motivations for diffusion distillation is to significantly accelerate the sampling speed, which is currently hindered by the large number of neural function evaluations required. To improve the inference efficiency of DMs, researchers are exploring ways to distill the learned knowledge from DMs to efficient sampling mechanisms, such as a direct implicit generator or a fewer-steps vector field. By doing so, they have been able to create student models that have further improved inference efficiency with minimal performance loss. Some distilled student models require less than 10 neural function evaluations but still offer comparable generative performance to their larger counterparts.

Diffusion distillation also serves as a means to establish connections between DMs and other generative models, such as implicit generative models and normalizing flows. Through knowledge transfer between DMs and other models, researchers can study the micro-connections between them and explore their potential for future generative modeling research.

This paper provides a comprehensive review of existing research on diffusion distillation strategies. Our review is organized into three main categories: diffusion-to-field (D2F) distillation (Section \ref{sec:d2f}), diffusion-to-generator (D2G) distillation (Section \ref{sec:d2g}), and training-free (TF) distillation (Section \ref{sec:asd}). Each category contains studies that share similar settings and methodologies. In addition to our categorization, we also discuss broader topics in diffusion distillation throughout the rest of this survey.

\section{Diffusion-to-Field Distillation}\label{sec:d2f}

The D2F distillation approach aims to address the lack of efficiency in the deterministic sampling method of DMs by distilling the generative ODE \eqref{def:rev_ode} into another generative vector field that requires fewer NFEs to generate comparable samples. This approach can be categorized into two classes: output distillation and path distillation. Output distillation aims to teach a student vector field to replicate the output of the DM's deterministic sampling method. Path distillation, on the other hand, aims to produce a student ODE that has better path properties than the teacher ODE. Both output and path distillations can be used in combination to improve both the path properties and simulation efficiency of the teacher ODE.

\subsection{Output Distillation}
To begin with, we first recap the generative ODE \eqref{def:rev_ode}
\begin{align*}
    dX_t = [F(X_t,t)-\frac{1}{2}G^2(t)\nabla_{X_t} \log p^{(t)}(X_t)]dt,~t\in [T,0], ~p^{(T)}=p_T
\end{align*}
To make the discussion simpler, we consider the most naive Euler-Maruyama (EM) discretization method that solves the ODE with sequential updates
\begin{align*}
    X_{t_i} = X_{t_{i+1}} + [F(X_{t_{i=1}},t_{i+1})-\frac{1}{2}G^2(t_{i+1})\nabla_{X_{t_{i+1}}} \log p^{(t_{i+1})}(X_{t_{i+1}})](t_{i}-t_{i+1}), ~i=N,...,1
\end{align*}
In the context of sampling from DMs, using EM methods directly for numerical solvers suffers from computational inefficiency. This is because the EM discretization error increases significantly with the step size, leading to poorly generated samples. To address this issue, an alternative approach called output distillation has been proposed. It involves training a student neural network to learn the output of the ODE using larger step sizes. Specifically, assuming that the step size $\Delta t$ is not too small, a student continuously-indexed neural mapping $S_\phi^{(stu)}(x,t)$ is trained to approximate the change in the ODE's output. This helps improve the computational efficiency of sampling from DMs by reducing the number of NFEs required for competitive sampling performance. More precisely, the student continuously-indexed neural mapping $S_\phi^{(stu)}(x,t)$ is trained to approximate the change of teacher ODE's output between time $t$ and $t-\Delta t$
\begin{align}\label{eqn:ode_residual}
    \Delta X \colon= X_{t-\Delta t} - X_{t} = \int_{s=t}^{t-\Delta t} [F(X_s,s)-\frac{1}{2}G^2(s)\nabla_{X_s} \log p^{(s)}(X_s)]ds.
\end{align}
The residual in \eqref{eqn:ode_residual} is tackled using numerical methods with sufficiently small step sizes. In order to overcome the computational inefficiency of this approach, a student network $S_\phi^{(stu)}$ is introduced as a distilled time-dependent vector field that approximates the non-linear residual in \eqref{eqn:ode_residual}. By leveraging the expressive power of neural networks, empirical studies have shown that properly designed diffusion-to-field techniques can result in a new sampling ODE that requires only one NFE but still yields comparable generative performance in terms of Fretchet Inception Distance (FID) \citep{}. It is worth noting that in the following sections, we use $\theta$ to denote the parameters of teacher models (DMs) and $\phi$ to represent those of student models unless otherwise specified.

\begin{figure}
\centering
\includegraphics[width=0.4\textwidth]{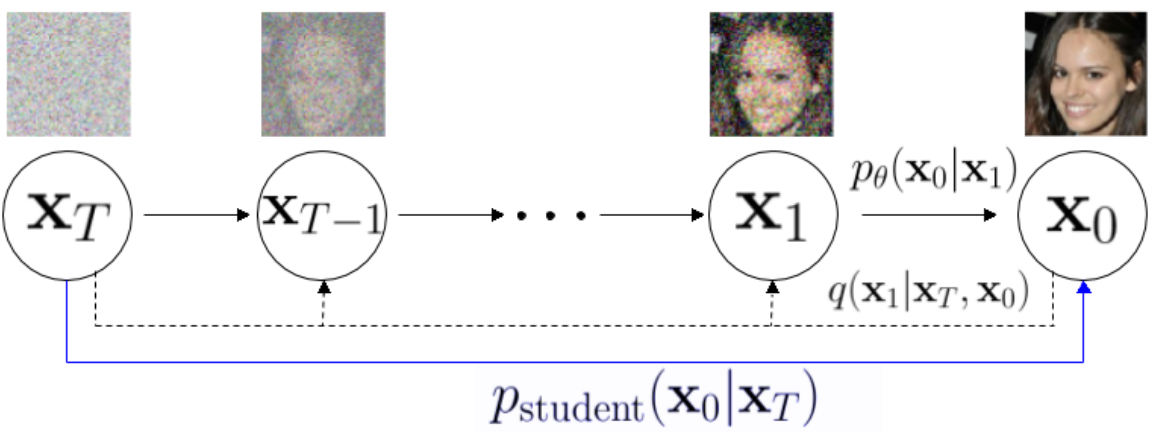}
\caption{Knowledge Distillation Strategy proposed in \citet{Luhman2021KnowledgeDI}.}
\label{fig:kd1}
\end{figure}

\citet{Luhman2021KnowledgeDI} propose a knowledge distillation (KD) strategy to distill a DDIM sampler to a Gaussian model that requires only one NFE when sampling. DDIM is a typical deterministic sampling strategy for VP diffusion models. Assume $\epsilon_\theta(x,t)$ is DM's noise-prediction network as discussed in \eqref{eqn:dsm_noise_predict}, the DDIM \citep{} sampler is a deterministic sampler that sequentially updates 
\begin{align}\label{eqn:ddim}
    X_{t_{i-1}} = \sqrt{\frac{\alpha_{t_{i-1}}}{\alpha_{t_i}}}(X_{t_i} - \sqrt{1-\alpha_{t_{i}}}\epsilon_\theta(X_{t_i})) + \sqrt{1-\alpha_{t_{i-1}}}\epsilon_\theta(X_{t_i})
\end{align}
DDIM is a deterministic sampling strategy that has been shown to significantly reduce the number of NFEs required for generating samples, with as few as 100 NFEs being sufficient for maintaining good generative performance. In contrast to the generative ODE method, DDIM is widely regarded as a superior deterministic sampling strategy. In their work, \citet{Luhman2021KnowledgeDI} propose to use a conditional Gaussian model as the student generative model
\begin{align}
    p_{stu}(x_0|x_T) = \mathcal{N}(x_0; f_\phi(x_T), \mathbf{I}).
\end{align}
The neural network $f_\phi(.)$ has the same input and output dimensions as the data. To implement it, the authors choose the architecture of $f_\phi$ to be the same as that of the score network in the teacher DM. Let $\operatorname{DDIM}(.)$ denote the deterministic mapping induced by the DDIM, they took
\begin{align*}
    p_{teacher}(x_0|x_t) = \mathcal{N}(x_0; 
\operatorname{DDIM}(x_T),\mathbf{I})
\end{align*}
To train the student model, they propose to minimize the conditional KL divergence between the student model and the DDIM sampler
\begin{align}
    \mathcal{L}(\phi) &= \mathbb{E}_{x_T\sim \mathcal{N}(0,\mathbf{I})}  \mathcal{D}_{KL}\bigg[p_{teacher}(x_0|x_T), p_{stu}(x_0|x_T)\bigg]\\
    &= \mathbb{E}_{x_T\sim \mathcal{N}(0,\mathbf{I})}  \mathcal{D}_{KL}\bigg[\frac{1}{2} \|f_\phi(x_T) - \operatorname{DDIM}(x_T)\|_2^2\bigg]
\end{align}
The student model sampling strategy proposed by \citet{Luhman2021KnowledgeDI} is simple. It involves drawing a Gaussian random variable $x_T$ from a normal distribution with mean 0 and identity covariance matrix. They then obtain the mean vector of the student model $x_0$ by passing $x_T$ through the neural network $f_\phi$. This results in a one-NFE sampling model with an FID of 9.39, while the teacher generative ODE has an FID of 4.16. Although this method provides a first step towards considering knowledge distillation for diffusion models, it has a computational inefficiency as it requires generating final outputs of DDIM or other ODE sampler, which consists of hundreds of NFEs for computing a single batch of training data. 

\begin{figure}
\centering
\includegraphics[width=0.4\textwidth]{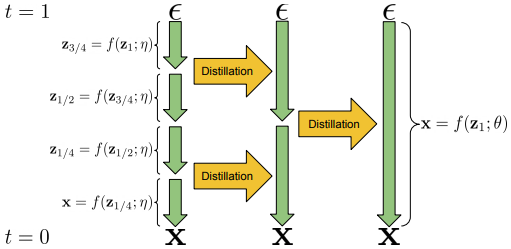}
\caption{Progressive Distillation Strategy proposed in \citet{Salimans2022ProgressiveDF}.}
\label{fig:kd1}
\end{figure}

The progressive distillation (PD) strategy proposed by~\citep{Salimans2022ProgressiveDF} aims to train a student neural network that requires half the number of NFEs as the teacher model by learning the two-step prediction of the teacher DM's deterministic sampling strategy. The teacher diffusion model is discretized to $N$ time-stamps, denoted by $\mathcal{T} = {t_i}{i=0, 1, ..., N-1}$, while $\mathcal{T}'= {0,2,4, ... }$ represents $N/2$ even time-stamps of $\mathcal{T}$. The student network is denoted by $f\phi(x,t)$, and the updates of the DDIM method from a time-stamp $t_j$ to another time-stamp $t_i$ in the teacher DM are denoted by $\operatorname{DDIM}(x,t_j,t_i)$ with $i\leq j$. The PD trains the student network by minimizing 
\begin{align}\label{eqn:pd_obj}
    \mathcal{L}(\phi) = \mathbb{E}_{x_0\sim p_d, i\sim Unif(\mathcal{T}'), \epsilon\sim \mathcal{N}(0,\mathbf{I})} \|f_\phi(\Tilde{x}, t_i) -\operatorname{DDIM}(\Tilde{x}, t_i, t_{i-2})\|,
\end{align}
where $\Tilde{x} = \sqrt{\alpha_{t_i}}x_0 + \sqrt{1-\alpha_{t_i}}\epsilon$ is the forward diffused data at time $t_i$ of VP diffusion. By minimizing the PD objective \eqref{eqn:pd_obj}, the student network learns to output a two-step prediction of the teacher model, so the total NFEs are halved. After the student model is trained to accurately predict the teacher model's two-step sampling strategy, it replaces the teacher model, and a new student model is trained to further reduce the number of sampling steps by half. The authors of the paper used the same UNet architecture as the teacher model's score network and the DDIM method as the initial teacher sampling strategy in their implementation of the progressive distillation (PD) approach. Their results showed that successive PD rounds can reduce the required vector fields to only 4 NFEs, making it 250 times more efficient than the teacher diffusion's ODE sampler, with a 5\% drop in generative performance as measured by FID. Both PD and KD use the teacher network's architecture to distill a few-step sampling method by learning from the many-step teacher sampling method. Both approaches are implemented by minimizing the $L^2$ error between the multi-step predictions of the teacher network and the single-step prediction of the student network. The key difference between PD and KD is that PD progressively reduces the number of required function evaluations, while KD directly trains a one-step student model for the final prediction. Thus, KD can be seen as an extreme PD method that reduces the teacher sampling strategies' full time-stamps to one in a single round.

A two-stage distillation strategy is proposed by \citet{Meng2022OnDO} to address the challenge of distilling knowledge from classifier-free guided conditional diffusion models like GLIDE \citep{Nichol2021GLIDETP}, DALL$\cdot$E-2 \citep{Ramesh2022HierarchicalTI}, Stable Diffusion \citep{Rombach2021HighResolutionIS} and Imagen \citep{Saharia2022PhotorealisticTD}. The key challenge is to transfer knowledge from the teacher DMs while preserving the classifier-free guidance mechanism, which successfully trains a single DM to learn both conditional and unconditional distributions. The conditional knowledge is integrated into the DM through a conditional input, while the unconditional knowledge is learned by replacing the conditional inputs with a \emph{None} input. In the first stage of their strategy, a student conditional diffusion model with a classifier-free guidance input is trained to learn from the teacher diffusion model. The student model is implemented as a neural network with learnable parameters, which takes in an input $x$ and a conditional context input $c$. For simplicity, the notation $c$ is dropped in the following discussion. The student model is trained to match the output by minimizing the objective 
\begin{align}
    \mathcal{L}(\phi_1) = \mathbb{E}_{w\sim p_w, t\sim U[0,1], x\sim p_d} \bigg[\lambda(t)\|f_{\phi_1}(x_t,t,w) - \hat{x}^w_\theta(x_t) \|_2^2 \bigg]
\end{align}
Here $\lambda(t)$ represents a pre-defined weighting function. $\hat{x}^w_\theta(x_t) = (1+w)\hat{x}_{c,\theta}(x_t) - w \hat{x}_{\theta}(x_t)$, $x_t\sim p_t(x_t|x_0)$ and $p_w = U[w_{min}, w_{max}]$. Note that the stage-one distillation only incorporates the classifier-free guidance through an additional input $w$ for the student model, no reduction of NFEs and efficiency-related distillation is applied. The second stage of distillation employs the progressive diffusion (PD) strategy proposed by \citet{Salimans2022ProgressiveDF} to significantly reduce the number of diffusion steps of the previously trained student model with classifier-free guidance inputs. The two-stage distillation approach is used to distill both pixel-space and latent-space classifier-free guided conditional diffusion models for various tasks, including class-conditional generation, text-guided image generation, text-guided image-to-image translation, and image inpainting. In fact, the distilled students achieved even better Fréchet Inception Distance (FID) scores than their teachers in the experiment of distilling pixel-space class-conditional generation on the ImageNet-64x64 dataset.

In recent work, \citet{Sun2022AcceleratingDS} proposed a feature space distillation method called Classifier-based Feature Distillation (CFD) for image data. The primary motivation behind this approach was to address the challenge of directly aligning pixels in the teacher's output and a few-step student model's output, which they found to be too difficult to learn. Instead, they trained the student networks to align with the multiple-step output of the teacher models in the feature space extracted by a pre-trained classifier. This approach follows a similar distillation strategy as the PD technique.

To achieve this, they proposed minimizing the KL divergence between the predicted probability distribution (after $\operatorname{Softmax}$ function) of the student model's one-step outputs and the teacher model's multiple-step outputs. They also found that incorporating additional terms such as entropy and diversity regularizations improved the distillation performance. They employed the same diffusion models and student models as \citet{Salimans2022ProgressiveDF} and used DenseNet-201 \citep{Huang2016DenselyCC} as the classifier in their implementations. Their distillation approach resulted in a student model with FID 3.80 on CIFAR10 with only 4 NFEs, which is lower than the DP's implementation in \citet{Salimans2022ProgressiveDF}. 

The PD strategy was also applied by \citet{Huang2022ProDiffPF} to distill fast text-to-speech (TTS) generation models based on a pre-trained diffusion TTS model. They introduced a variance predictor and a spectrogram denoiser to improve the student model's architecture, which was specifically designed for TTS applications.

In contrast to mimicking the output of the generative ODE of diffusion models, \citet{song23consistency} proposes to minimize the difference of self-consistency function of generative ODE to implement the output distillation. They randomly diffuse a real data sample and simulate a few steps of generative ODE to obtain another noisy data which lies on the same ODE path. They input the two noisy samples into a student model and minimize the difference in the outputs in order to ensure the self-consistency of generative ODE. They name their proposed model the consistency model (CM). CM can be viewed as another output distillation method that utilizes the self-consistent property of generative ODE for distillation.

For output distillation, the student network is trained to minimize the difference between its output and the output of the teacher ODE at the same time points. The output distillation is particularly useful when the teacher's ODE has a relatively simple form, such as diagonal ODEs, or when the teacher's ODE is difficult to train. In these cases, the student network can be trained to efficiently mimic the teacher ODE's output with much fewer NFEs. The trained student network can then be used for efficient sampling from the DMs.

It is worth noting that the output distillation only considers the output of the ODE at discrete time points, and thus may not fully capture the path properties of the DMs. Therefore, it may not be suitable for applications where path properties are important, such as image synthesis and editing. For such applications, path distillation can be used instead, which will be discussed in the next section.

\subsection{Path Distillation} 
Output distillation is a technique that helps improve the sampling strategy of DMs by allowing the student neural network to mimic the multiple-step output of teacher models. In contrast, path distillation aims to refine DMs' sampling strategy to potentially have better properties. Some researchers argue that the forward (and reverse) diffusion process creates a curve in data space that connects the data distribution and prior distribution. Thus, path distillation focuses on refining the diffusion generative SDE or ODE to a straight version that has a more efficient sampling path than the teacher models. The key difference between path distillation and output distillation is that path distillation is more concerned with refining an existing teacher model's sampling strategy, whereas output distillation focuses on teaching the student to learn the skipped output of the teacher model, without changing the sampling path and mechanism of the student model.

The Reflow method, introduced by \citet{Liu2022FlowSA}, is a path distillation approach that aims to improve generative speed by modifying a pre-trained teacher neural ODE through a student model. The student model straightens the path of the teacher model by minimizing a time average of $L^2$ loss between its outputs and interpolations of data samples and corresponding outputs from the pre-trained model. More precisely, let $p_T$ be an initial distribution
\begin{align}
    \mathcal{L}(\phi) = \mathbb{E}_{x_T\sim p_T, x_0\sim p_{teacher}(x_0|x_T), t\sim U[0,T]} \bigg[ \|x_T - x_0 - f_\phi(\frac{t}{T} x_T + \frac{T-t}{T} x_0, t)\|_2^2 \bigg]
\end{align}
The term $p_{teacher}(x_0|x_T)$ can be an arbitrary teacher model regardless it is an SDE or ODE. In Reflow distillation proposed by \citet{Liu2022FlowSA}, if the teacher model is assumed to be a DDIM sampler, then $p_{teacher}(x_0|x_T) = \delta(x_0=\operatorname{DDIM}(x_T))$. Reflow distillation has the advantage of being repeatable for several rounds, which can further straighten the teacher model's path. Additionally, Reflow can be used in conjunction with progressive distillation as discussed in Section \ref{}, where the Reflow strategy can be used first to straighten the teacher model's path, followed by progressive distillation to make the student model more efficient. Finally, the authors provided numerical comparisons between ODE paths of rectified and un-rectified models.

\begin{figure}
\centering
\includegraphics[width=0.7\textwidth]{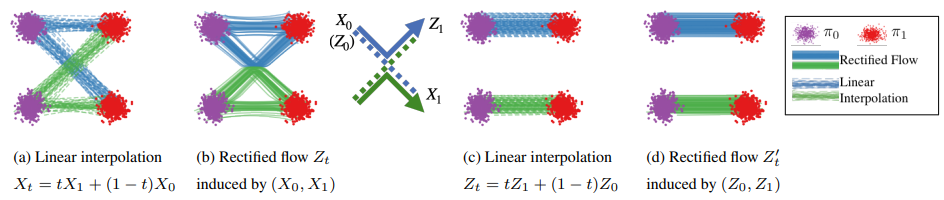}
\caption{Reflow Strategy for path distillation. The figure is taken from \citet{Liu2022FlowSA}.}
\label{fig:reflow1}
\end{figure}

\begin{figure}
\centering
\includegraphics[width=0.7\textwidth]{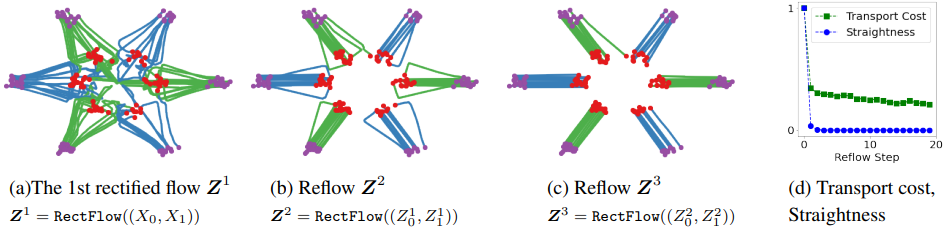}
\caption{Reflow Strategy for path distillation. The figure is taken from \citet{Liu2022FlowSA}.}
\label{fig:reflow2}
\end{figure}

In their work, \citet{Wu2022FastPC} applied the Reflow technique to the problem of 3D point cloud generation. They proposed a three-stage procedure that enables the construction of a generative ODE capable of producing high-quality point clouds. At their first training stage, they train a teacher generative ODE $f_\theta$ by minimizing the objective
\begin{align}
    \mathcal{L}(\theta) = \mathbb{E}_{x_0\sim p_d, x_T\sim \mathcal{N}(0,\mathbf{I}), t\sim U[0,T]} \bigg[ \|f_\theta(x_t,t) - (x_0-x_T)\|^2_2 \bigg]
\end{align}
where $x_t = (t/T)x_0 + (T-t/T)x_T$ is the interpolated point at time $t$. In the second stage, they applied the Reflow strategy to further \emph{straighten} the teacher model. In the third stage, they use a student model $f_\phi$ which finally distills the multiple-step teacher model into a single-step student model by minimizing 
\begin{align}
    \mathcal{L}(\phi) = \mathbb{E}_{x_T\sim p_d} \operatorname{Dist}\bigg[x_T + f_\phi(x_T,T), x_0\bigg]
\end{align}
Here $x_0 \sim p_{teacher}(x_0|x_T)$ is obtained from teacher model. The term $\operatorname{Dist}(.,.)$ represents some distance function between two points. In their implementation, they use the Chamfer distance for measuring the distance between two cloud points. 

\citet{Lee2023MinimizingTC} proposed an objective that can result in a forward process with much smaller curvations, to be more insensitive to truncation errors. \citet{Chen2023RiemannianFM}, \citet{Li2023SelfConsistentVM} and \citet{Albergo2022BuildingNF} further study the refinements of the diffusion model's generative path. \citet{Zheng2022FastSO} proposed another view for path distillation, which learns a mapping operator which could generate the path. \citet{Fan2023OptimizingDS} also proposes another path distillation method. They refine the generative path by finetuning the teacher ODE to minimize some IPM according to the forward path. \citet{Aiello2023FastII} proposed to finetune the generative path by minimizing the MMD between each marginal distribution of the generative path and the data. 

The diffusion-to-field distillation remains an active area of research in the field of efficient diffusion models. The aim is to use the knowledge gained from diffusion models and generative SDE/ODE to distill a student model with a faster sampling speed while maintaining a comparable level of generative performance to that of the teacher models.

\section{Diffusion-to-generator Distillation}\label{sec:d2g}
In contrast to the diffusion-to-field distillation discussed earlier, diffusion-to-generator (D2G) distillation is another important category of distillation methods. The primary goal of D2G distillation is to transfer the distributional knowledge learned by the diffusion model into an efficient generator. Unlike D2F distillation, which focuses on learning student models with the same input and output dimensions, D2G distillation usually involves implicit generators that may not have the same latent dimension as the data space. Additionally, both deterministic and stochastic generators are considered as student generators, depending on the specific application. The training objective for D2G distillation typically takes a similar form as the diffusion model's training objective, as opposed to the commonly used mean square error-based objective in D2F distillation.

\subsection{Distill Deterministic Generator} 
There is increasing attention on distilling deterministic generators (e.g. neural radius field) as student models for further applications of pre-trained large-scale diffusion models. More specifically, the pre-trained text-to-image diffusion models are found notably useful for learning neural radius fields which have contents that are related to some given text prompts. The neural radius field (NeRF)~\citep{Mildenhall2020NeRFRS} is a kind of 3D object which use a multi-layer-perceptron (MLP) to map coordinates of a mesh grid to volume properties such as color and density. Given the camera parameters such as the angles of the views, the rendering algorithm outputs a 2D image that is a view projection of the 3D NeRF scene. From a given view, the NeRF is viewed as a deterministic 2D image generator whose MLP's parameters are learnable. 

\begin{figure}
\centering
\includegraphics[width=0.7\textwidth]{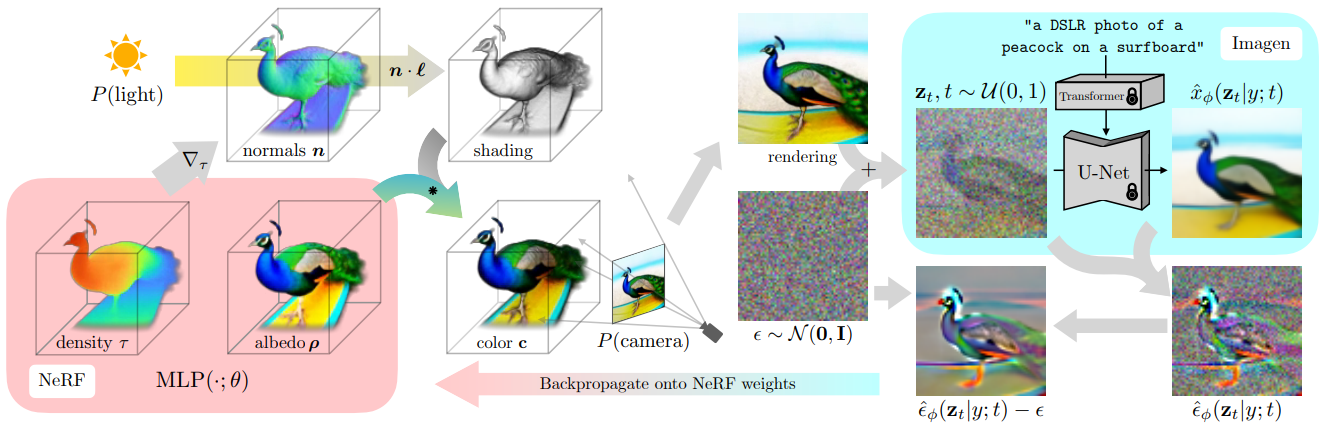}
\caption{Score Distillation Sampling method. The figure is taken from \citet{poole2022dreamfusion}.}
\label{fig:sds1}
\end{figure}

The limited availability of data for constructing NeRFs has motivated researchers to explore distillation methods to obtain NeRFs with contents related to given text prompts. The pioneering work by Poole et al.~\citep{Poole2022DreamFusionTU} proposed a method called score distillation sampling (SDS) to distill a 2D text-to-image diffusion model into 3D NeRFs. Unlike traditional NeRF construction that requires images from multiple views of the target 3D objects, text-driven construction of NeRF lacks both the 3D object and the multiple views. The SDS method optimizes the NeRF by minimizing the diffusion model's loss function using NeRF-rendered images from a fixed view. To avoid the computational expense of directly optimizing the diffusion model's loss function, the researchers propose to approximate the distillation objective by omitting the Unet jacobian term. Specifically, the rendered NeRF image from a fixed view with the learnable parameter $\phi$ is represented by $x=g(\phi)$, and the forward diffusion sample is denoted as $x_t = \sqrt{\alpha_t}x_0 + \sqrt{1-\alpha_t}\epsilon$ as proposed in \eqref{eqn:vp_reparam}. The trained text-conditional diffusion model is denoted by $\epsilon_\theta(x,t,c)$, where $t$ and $c$ represent the time-stamp and text prompt. The SDS uses the gradient to update parameter $\phi$ with 
\begin{align}
    \operatorname{Grad}(\phi) = \frac{\partial}{\partial \phi} \mathcal{L}(\phi) = \mathbb{E}_{t,\epsilon,x=g(\phi)} \bigg[w(t)(\epsilon_\theta(x_t,t,c) - \epsilon)\frac{\partial x}{\partial \phi} \bigg]
\end{align}
The work by \citet{Poole2022DreamFusionTU} achieved remarkable results in the task of generating 3D NeRFs from text prompts, which has spurred further research on the distillation of diffusion models into deterministic generators.

Several works have extended the SDS approach to other applications, as reported in \citet{Wang2022ScoreJC,Lin2022Magic3DHT,Deng2022NeRDiSN,Xu2022NeuralLift360LA,Singer2023TextTo4DDS}. In particular, \citet{Wang2022ScoreJC} has conducted a thorough investigation of the tunable parameters of the SDS algorithm for text-to-3D generation. Meanwhile, \citet{Lin2022Magic3DHT} have proposed a two-stage optimization strategy that further enhances the performance of SDS on high-resolution images. The first stage of their approach is similar to that of \citet{Poole2022DreamFusionTU}, which obtains a low-resolution NeRF. The second stage up-scales the trained NeRF from the first stage to a higher resolution and fine-tunes it for better performance.

The successful application of distillation to deterministic generators, particularly in the domain of 3D generative modeling, has been widely acknowledged. However, the research in developing better-performing and more efficient distillation strategies is still largely unexplored, making it a hot research topic that demands further exploration.

\subsection{Distill Stochastic Generator}
In this section, we will discuss distillation strategies for stochastic generators, also known as implicit generative models. These models have been widely used in generative modeling and differ from deterministic generators in that they use neural transforms to map latent vectors to generate data samples with randomness. Stochastic generators have shown great success in the past decade \citep{Goodfellow2014GenerativeAN, Karras2019AnalyzingAI, Brock2018LargeSG}, with their advantages being fast inference speed, low inference memory, and lightweight models. Distilling diffusion models into stochastic generators has been motivated by the need for extreme inference efficiency.

The work by \citet{Luhman2021KnowledgeDI} can be interpreted as a type of score distillation for stochastic generators. Specifically, they employ a Unit, which has the same neural architecture as the pre-trained diffusion model, to map a latent vector and calculate the mean of a Gaussian output. The dimensionality of their latent vector matches that of the output data, and they assume unit variance for Gaussian conditional generation. To train the student model, they minimize the conditional KL divergence between the generator's output and a deterministic sampler for the diffusion model. Since they use conditional Gaussian output, minimizing KL divergence is equivalent to minimizing the mean square error between the model's mean and the sampler's output, as discussed in Section \ref{sec:d2f}.

\section{Accelerated Sampling Algorithms as Diffusion Distillation}\label{sec:asd}
The diffusion model has the unique characteristic of separating training and sampling processes. During training, the diffusion model does not require sampling, and during sampling, there is a high level of flexibility to design and improve the process. Typically, the diffusion model is trained with all discrete or continuous noise levels, but sampling from it does not necessarily require querying all diffusion time levels. Recent research has shown that using a small subset of diffusion time levels can significantly accelerate the sampling process with much fewer NFEs. These accelerated sampling algorithms can be viewed as generalized distillations of diffusion models, where the goal is to train a big model but use a small one. We can further categorize these algorithms into training-free and training-based accelerating algorithms as two distinct categories of diffusion distillation.

To train a diffusion model, a neural score network is trained to match the marginal score functions of a data diffusion process. As \citet{Song2020ScoreBasedGM} pointed out, the reversed SDE \eqref{def:rev_sde} and ODE \eqref{def:rev_ode} share the same marginal distribution as the forward diffusion and can be used for sampling if simulated backward. Therefore, the reversed ODE or SDE serves as the starting point for sampling from diffusion models. Formally, the most naive simulation of reversed SDE uses an Euler-Maruyama discretization with a formula
\begin{align}\label{eqn:rev_sde_em}
    X_{t+1} = X_t - \big[ F(X_t,t)-G^2(t) S_\theta(X_t,t) \big] \Delta t + G(t)\Delta t, t= T,...,1.
\end{align}
Here $S_\theta(x,t)$ represents the trained score network, i.e. the teacher model. The simple sampler presented in equation \eqref{eqn:rev_sde_em} updates a batch of samples by simulating the reversed SDE, taking all diffusion time levels into account sequentially. However, this sampling strategy that uses all-time levels suffers from poor computational efficiency. In later sections, we will introduce both training-based and training-free accelerated sampling algorithms. These algorithms aim to improve the sampling speed of diffusion models by utilizing a smaller subset of diffusion time levels.

\subsection{Training-based Accelerating Algorithms}
The selection of appropriate diffusion time levels is a crucial problem for accelerating sampling algorithms.

Previous works have shown that it is possible to construct more efficient samplers that have comparable performance to the naive sampler by using only a subset of diffusion time levels. \citet{Watson2022LearningFS} addressed this problem by proposing to solve a dynamic programming problem with the Fréchet Inception Distance (FID) metric as the target. They learned a scheduler model to choose the subset of steps to be used in the sampling process, resulting in better efficiency.

To improve the generative performance of diffusion models, \citet{Bao2022EstimatingTO} proposed to train additional covariance networks. The idea is to capture the full covariance matrix of the latent space distribution, instead of assuming diagonal covariance as in the original diffusion model. By doing so, the model can better capture the complex correlation structure of the data and generate more realistic samples. The authors achieved promising results on various datasets and showed that their method outperforms the baseline diffusion model with diagonal covariance.

The approach proposed by \citet{Kim2022RefiningGP} involves using a likeliratio-estimation model, which is essentially a discriminator, to learn the difference between the ground truth score function and the model's score function. This difference is then used to obtain an unbiased score function by combining it with DM's score functions, after differentiating through the learned diffusion. The resulting refined DM is expected to offer improved generative performance even with fewer NFEs.

\subsection{Training-free Accelerating Algorithms}
The training-free algorithms aim to achieve comparable generative performance with a much fewer number of used diffusion time levels. The most distinguished part of training-free accelerating algorithms is to design faster sampling algorithms with no training of new parametric models but only inferencing by querying pre-trained DMs. Most of such algorithms center around different numerical solvers of generative SDE or ODE. 

To begin with, we start with the denoising diffusion probabilistic models (DDPM) \citep{Ho2020DenoisingDP}. The DDPM implements a discretized version of VP diffusion and uses a discretized sampling scheme with
\begin{align}
    x_{t_{i-1}} = \frac{1}{\alpha_{t_i}}(x_{t_i}-\frac{1-\alpha_{t_i}}{\sqrt{1-\Bar{\alpha}_{t_i}}}\epsilon_\theta(x_{t_i},t_{i})) + \sigma_{t_i} z_i
\end{align}
Here $\alpha_{t_i} = 1-\beta_{t_i}, \Bar{\alpha}_{t_i} = \prod_{j=1}^i \alpha_{t_i}\approx e^{-\int_{s=0}^{t_i} e^{\beta(s)}ds}$ is a discretized inplementation of integral. $\sigma_{t_i}$ is an arbitrary $\sigma$ levels and $z_i$ is an independent Gaussian noise. The DDPM uses all $1000$ diffusion levels (i.e. $\{t_1,...,t_{1000}\}$) for sampling. 

\citet{Song2020DenoisingDI} re-formulate the derivation of DDPM under a non-Markov forward diffusion model and results in a new family of sampler
\begin{align}
    x_{t_{i-1}} = \sqrt{\alpha_{t_{i-1}}}\bigg[ \frac{x_{t_i} - \sqrt{1-\alpha_{t_i}}\epsilon_\theta(x_{t_i}, t_i)  }{\sqrt{\alpha_{t_i}}} \bigg] + \sqrt{1-\alpha_{t_{i-1}} - \sigma_{t_i}^2}\cdot \epsilon_\theta(x_{t_i}, t_i) + \sigma_{t_i} z_{t_i}
\end{align}
The term $\alpha_{t_i}$ has the same meaning as the one in DDPM's sampling algorithm. The $\sigma_{t_{i}}$ is a free hyper-parameter that controls the strength of randomness in the DDIM sampler. When $\sigma=\sqrt{(1-\alpha_{t_{i-1}})/(1-\alpha_{t_i})}\sqrt{1-\alpha_{t_i}/\alpha_{t_{i-1}}}$, the DDIM sampler coincides with the DDPM sampler. The sampler is deterministic when $\sigma_{t_i} = 0$. The DDIM sampler requires the same pre-trained DM as the DDPM's sampler and is shown to be able to remain the generative performance by querying fewer diffusion time levels for sampling. 

The work of \citet{Bao2022AnalyticDPMAA} showed that there is an optimal reverse variance $\sigma_{t_i}$ in the DDIM sampler family that minimizes the KL divergence between the forward and reversed Markov chain of DDIM. Moreover, they derived an explicit expression for this optimal variance with the form 
\begin{align}
    \sigma_n^{*2} = \lambda_n^2 + \bigg[\sqrt{\frac{\Bar{\beta}_n}{\alpha_n}} - \sqrt{\Bar{\beta}_{n-1} - \lambda_n^2}\bigg]^2\cdot \bigg[1-\Bar{\beta}_n \mathbb{E}_{q_n(x_n)}\frac{\|\nabla_{x_n}\log q_n(x_n)\|^2}{d} ) \bigg]
\end{align}
They claimed the DDIM sampler with such optimal variance leads to improved sampling regardless of the pre-trained diffusion models. In practice, the optimal variance is estimated with the Monte Carlo method with the help of pre-trained DMs, with the form
\begin{align}
    \Gamma_n = \frac{1}{M} \sum_{m=1}^M \frac{\|s_n(x_{n,m})\|^2}{d},x_{n,m}\sim^{iid} q_n(x_n)
\end{align}
\begin{align}
    \hat{\sigma}^2_n = \lambda_n^2 + \bigg[ \sqrt{\frac{\Bar{\beta}_n}{\alpha_n}} - \sqrt{\Bar{\beta}_{n-1} - \lambda^2_n} \bigg]^2 (1-\Bar{\beta}_n \Gamma_n).
\end{align}

\citet{Liu2022PseudoNM},\citet{Zhang2022FastSO} and \citet{Lu2022DPMSolverAF} find out the semi-linear structure of VP's reversed ODE
\begin{align}
    \frac{dX_t}{dt} = F(t)X_t - \frac{1}{2} G^2(t)\nabla_{X_t} \log p^{(t)}(X_t)
\end{align}
where $F(t) = -\frac{1}{2}\beta(t)$ and $G(t)=\sqrt{\beta(t)}$ for VP diffusion. They applied an exponential integrator technique on ODE that further simplified VP's reversed ODE as
\begin{align}
    x_t = \frac{\alpha_t}{\alpha_s}X_s - \alpha_t \int_s^t \frac{d\lambda_\tau}{d\tau} \frac{\sigma_\tau}{\alpha_\tau}\epsilon_\theta(X_\tau, \tau)d\tau
\end{align}
Where $\lambda_t\colon= \log (\alpha_t/\sigma_t)$ is the log-SNR function. \citet{Lu2022DPMSolverAF} further proposed a change of variable trick and Taylor expansion on a simplified algorithm and obtain higher-order solvers for VP reversed ODE. 

Higher order SDE solvers have been proposed by \citet{JolicoeurMartineau2021GottaGF} as an alternative to EM discretization, and they have demonstrated improved sampling performance in terms of FID. Building on this work, \citet{Karras2022ElucidatingTD} have suggested better neural preconditioning of DMs along with the use of second-order Heun discretization for simulating reversed ODE and SDE. Their method has set a new record for generative performance for DMs with a score of 1.79 FID.

Other research works have also explored the use of advanced numerical solvers for DMs. For example, \citet{Li2023ERASolverEA}, \citet{Wizadwongsa2023AcceleratingGD}, and \citet{Lu2022DPMSolverFS} have investigated the potential of different numerical solvers to accelerate DM's inference efficiency. The key idea of this training-free accelerated sampling algorithm is to achieve faster sampling without the need for additional training of parametric models. However, even with the best-performing training-free methods, more than 10 NFEs are typically required to achieve comparable generative performance. \citet{zhao2023unipc} proposes a unified framework that introduces the prediction-correction-type numerical solutions to analyze and design higher-order ODE solvers for diffusion models. Their proposed UniPC framework includes many well-studied solvers as their special cases.

\section{Conclusion}
This paper provides a comprehensive review of existing techniques for knowledge distillation of diffusion models. Our review is organized into three categories: diffusion-to-field distillation, diffusion-to-generator distillation, and accelerating algorithms as diffusion distillation. For each category, we discuss several landmark methodologies and their applications. Given the growing popularity of diffusion models in recent years, knowledge distillation of diffusion models has become an increasingly important research area that can impact the efficient application of these models. Although significant progress has been made in these areas, there is still much room for improvement in the efficiency and effectiveness of knowledge distillation of diffusion models.

\newpage
\bibliography{reference} 

\appendix

\end{document}